# Probabilistic Medical Predictions of Large Language Models

Bowen Gu[1], Rishi J. Desai[1], Kueiyu Joshua Lin[1*], Jie Yang[1,2,3*]

[1] Division of Pharmacoepidemiology and Pharmacoeconomics, Department of Medicine, Brigham and Women's Hospital, Harvard Medical School, Boston, MA, USA.

[2] Harvard Data Science Initiative, Harvard University, Cambridge, MA, USA

[3] Broad Institute of MIT and Harvard, Cambridge, MA, USA

**Address for Correspondence:**
Jie Yang, PhD
Assistant Professor of Medicine,
Division of Pharmacoepidemiology and Pharmacoeconomics, Department of Medicine, Brigham and Women's Hospital, Harvard Medical School
1620 Tremont Street, Suite 3030R Boston, MA 02120, USA
Phone: 857-260-3191 | Email: jyang66@bwh.harvard.edu
* Co-senior authors (J.Y and K.J.L contributed to the work equally).

**ABSTRACT**

Large Language Models (LLMs) have shown promise in clinical applications through prompt engineering, allowing flexible clinical predictions. However, they struggle to produce reliable prediction probabilities, which are crucial for transparency and decision-making. While explicit prompts can lead LLMs to generate probability estimates, their numerical reasoning limitations raise concerns about reliability. We compared explicit probabilities from text generation to implicit probabilities derived from the likelihood of predicting the correct label token. Across six advanced open-source LLMs and five medical datasets, explicit probabilities consistently underperformed implicit probabilities in discrimination, precision, and recall. This discrepancy is more pronounced with smaller LLMs and imbalanced datasets, highlighting the need for cautious interpretation, improved probability estimation methods, and further research for clinical use of LLMs.

**INTRODUCTION**

Generating credible probability of prediction is crucial in clinical practice and medical research when applying artificial intelligence (AI) to healthcare. Reliable probability outputs are critical for informed decision-making, stratifying patients with different risk levels, and enabling clinicians to set probability thresholds according to their preferred trade-offs between recall and precision in real-world clinical practice[1-4]. For example, a lower threshold can be applied in screening applications to reflect the aversion to false negatives. Accurate probability outputs have a strong influence the adoption and effectiveness of AI in healthcare. Discriminative AI models, such as support vector machines and classic deep learning models[5,6], predict labels by assigning probabilities to a fixed set of label candidates and selecting the label with the highest probability. This paradigm naturally generates the probability of the predicted label and has been widely adopted in medical AI applications[5,7-9].

Generative AI models, especially Large Language Models (LLMs), have demonstrated remarkable general-purpose capabilities and the ability to perform few-shot or zero-shot learning, enabling accurate predictions with little or no annotated data[10,11]. These models are particularly advantageous in clinical applications where tasks are diverse and annotated data is scarce and expensive to generate[12,13]. Prompt-based LLMs represent the most common usage of LLMs, as they can flexibly instruct the models to perform different tasks by simply defining prompt text without heavy model training[14-16]. However, this generative framework does not naturally output probabilities of the predictions, as it converts all the tasks as a text generation process, rather than assigning probabilities to fixed candidate labels. As a result, probabilistic predictions of LLMs in healthcare have been reported and evaluated rarely[17-19], leading to a lack of crucial assessment of AI in clinical practice.

Probabilities have been estimated in the literature directly through text generation by LLMs. Such a probability, called explicit, is simple and flexible[20,21]. For example, one can instruct the LLM output the probability of its prediction by adding a sentence such as "Please provide the probability along with your prediction" to the prompt. The simplicity of explicit probability ensures that it can be applied in any advanced prompt engineering techniques such as Chain-of-Thought (CoT)[22], retrieval augmented generation (RAG)[23,24], self consistency[25], and LLM agents[26,27]. However, given the challenges LLMs face with numerical reasoning[28,29], the reliability of text-generated explicit probability values may be questionable and has not yet been thoroughly examined.

On the other hand, several works have delved into the LLM-based methods of probability prediction[21,30], which we define as "implicit probability". For example, for the question "*Given the following lab reports: [lab report text]. Does the patient have COVID? Your answer must be either 'Yes' or 'No'.*", and if the LLM response is "No", one can extract the probability of each generated token of the LLM, identify the position of prediction token (i.e. "No" in this example), and assign the model's implicit probability of the prediction as the probability that corresponds to the prediction token. However, such extraction of implicit probability is only available in limited scenarios. As shown in **Figure 3a**, in information extraction tasks, the LLM predicted label tokens are diverse and with different lengths; when advanced LLM prompts, such as CoT, are applied, the predicted label tokens are not in a fixed position. These factors complicate the extraction of implicit probability, making it challenging to automate the extraction across different tasks and prompts. In addition, many advanced proprietary LLMs, such as Google Gemini (https://deepmind.google/technologies/gemini/) and Anthropic Claude (https://claude.ai/), do not provide APIs that would return the probability for each output token, which makes it infeasible to get the implicit probability. Consequently, this implicit probability can only be extracted under very simple settings (e.g., for option selection tasks or tasks without a chain of thought) and is mostly available in open-source LLMs, which largely limits its application.

In this study, we extensively examine the reliability, which is defined as the quality of being trustworthy, of the probabilistic output of LLMs by comparing the explicit probability and the implicit probability. Specifically, we select 6 advanced open-weighted LLMs from different organizations and evaluated their performance on 5 medical datasets. To utilize the Area Under the Receiver Operating Characteristic (AUROC) and Area Under the Precision Recall Curve (AUPRC) as evaluation metrics, we convert the LLM prediction task into a binary selection task within a question-and-answer setting with two possible options. We find that while explicit probability reflects a certain level of model prediction confidence, its reliability is consistently lower than that of implicit probability across all LLM models and datasets, especially in the case of small LLMs and imbalanced datasets.

The contributions of this work include a large-scale evaluation of the probabilistic medical predictions of LLMs, a crucial component of applying AI in healthcare, demonstrating that the commonly used explicit probability of LLMs has lower reliability under certain circumstances, which necessitates caution when applying it in the medical domain, and providing a framework for evaluating the reliability of probabilistic outputs of LLMs in medical predictions, which can be easily extended to other fields requiring high-quality probabilistic assessments.

**RESULTS**

All LLMs follow instructions well, with outputs strictly adhering to the format in the prompt, except for the Qwen2-7B model on the USMLE dataset. All other models have a less than 10% fail rate on the validation process, with 78.6% of the model inference on one dataset have a less than 1% fail rate on the validation process. The details of instruction adherence for each LLM experiment are shown in **Supplementary Table 7**.

**LLM Performance**

The accuracy, AUROC, and AUPRC of the large LLMs on different datasets are listed in **Table 1** to **Table 3**, respectively. The "Explicit" stands for the AUROC of the explicit probability, "Implicit" stands for the AUROC of the implicit probability, and "Difference" stands for the difference between the AUROC of the implicit probability and the AUROC of the explicit probability. For **Table 2** and **Table 3**, the "P-value" indicates the statistical significance of the implicit probability over the explicit probability. According to **Table 1**, all LLMs achieve at least a 0.8 average accuracy on all datasets. Specifically, the Llama-3.1-70B model has the highest average accuracy among all the LLMs tested. The Qwen2-72B model, which is trained using a rich Chinese corpus, shows a clear advantage over other models on the MCMLE dataset, which consists of Chinese medical examination questions. On the other hand, the Phi-3-medium model has a poor performance on the MCMLE dataset. The Mistral-Large model, which has comparable capability to the Llama-3.1-70B model, achieve the highest accuracy for the MMLU-CM and the MGB-SDoH dataset but its performance is much worse than the Llama-3.1-70B model on the non-English MCMLE dataset (85.2% VS 93.0%).

Although the AUROC and AUPRC for both the explicit and implicit probabilities are high, the metrics for implicit probabilities are consistently higher than those for explicit probabilities across all LLMs, and the majority of the results show that such advantage is statistically significant. This finding is valid on both a non-English dataset (MCMLE) and a dataset from private electronic health records (MGB-SDoH). The only exception is the Llama-3.1-70B model on the USMLE dataset, which is mainly due to the high standard deviation of the implicit probability. Overall, this result indicates that for LLMs, the implicit probability is a better indicator of the model answer's confidence (**Tables 2** and **3**).

The ROC and PRC curves of the LLMs on each dataset are shown from **Supplementary Figure 1** to **Supplementary Figure 10**. We observe that the ROC and PRC curves for the explicit probability is a coarse step function because the LLMs usually give their probabilities in their nearest tenth (e.g. 90%, 80%, etc.), which makes the explicit probability coarse-grained. In contrast, the steps of the implicit probability are barely discernible in the plot. This further proves that the implicit probability is a better indicator than the explicit probability.

We find that in some cases, while the LLM provides an answer and an explicit probability as instructed, this probability is less than 50% (**Supplementary Table 3**). This implies that although the model proposes an answer, it assigns a higher probability to the alternative, undermining its reliability . In contrast, the implicit probability always selects the most probable token, avoiding such contradictions and supporting its reliability over the explicit probability.

**Large LLM vs. Small LLM**

The AUROCs of the large LLMs and their smaller counterparts on the USMLE and MGB-SDoH datasets are shown in **Figure 1**. Except for the Yi-1.5 model on the MGB-SDoH dataset, all large LLMs outperform their smaller versions on both explicit and implicit probability AUROC, due to their richer knowledge and stronger reasoning abilities. Furthermore, small LLMs generally exhibit greater differences between explicit and implicit probability AUROC, as their limited parameters make them less sensitive to explicit probabilities, reducing reliability. In contrast, the implicit probabilities difference for small LLMs is much smaller than for the explicit probabilities, indicating their better credible to represent the model's actual probability to its prediction.

**Imbalanced Datasets Analysis**

The AUPRC differences of the large LLMs are shown in **Figure 2a**, where the percentage of the imbalance in the legend indicates the percentage of option A being the correct label in the dataset, and a 50% imbalance means that there is an equal number of option A and option B being correct in the dataset. The difference in the figure is defined as the difference between the AUPRC of the implicit probability and the explicit probability. Based on **Figure 2**, for most LLMs the AUPRC difference increases as the dataset got more imbalanced. This can be explained by how the AUPRC is calculated. Since we define option A as positive when plotting the AUPRC, the scarcity of option A being the correct label plays an important role in determining the precision and the recall, which are the keys to the PRC curve. As the number of options A being correct gets lower, any false classification will be magnified, which enlarges the differences between the two probabilities if one is better than the other. We do not show the results under the 70% and 90% imbalance rates since they are very similar to the ones under the 50% imbalance ratio. This is also reasonable given that the PRC is not related to the true negatives, and the excessive number of options A being correct will minimize the differences between the two probabilities.

**LLM probability distribution**

The probability distribution of each LLM on each dataset is shown from **Supplementary Figure 11** to **Supplementary Figure 15** and an example of the Meta-Llama-3.1-70B-Instruct LLM on the USMLE dataset is shown in **Figure 2b**. According to **Figure 2b**, except for the Phi-3-medium model on the MCMLE dataset, all other LLMs have a relatively great performance on each dataset by having low probability when the true label is 0 and high probability when the true label is 1, which is consistent with the consistent performance we find (**Table 1**). Compared to the explicit probability, the implicit probability has a more dispersed distribution, in accord with our finding on the AUROC and AUPRC curves. However, we notice that the explicit probability, even with its narrow distribution, may give values that are close to 50%, while the more widely distributed implicit probability values rarely fall within this area.

For most LLMs, both their explicit and implicit probability distributions are very polarized, even if their predictions are incorrect. This indicates that the LLMs are overly confident about most of their predictions, regardless of their actual correctness. Since such polarization still persists on the MGB-SDoH dataset, a private dataset that can't be touched during LLM training, we argue that such confidence does not originate from the data leakage[31], but is an intrinsic property of the LLMs. Some exceptions to the above observation are the Mistral-Large and the Phi-3-medium model on the MCMLE and the MGB-SDoH dataset, where models show a broad spectrum of implicit probability distribution, but their explicit probability distributions are still highly polarized.

**Sensitivity Analysis**

The AUROCs of the LLMs on the USMLE dataset using the three different prompts are shown in **Supplementary Figure 16**. It indicates that the implicit probability has a higher AUROC than that of the explicit probability under all three prompts. We also learn that the main prompt and the first auxiliary prompt have similar performance in terms of AUROC for both probabilities, which is reasonable since both prompts use a similar multiple-choice format and contain similar information. The AUROC for both probabilities largely decrease when using the second auxiliary prompt. We propose that this is due to the introduction of the setting of a student proposing answers to the LLM, which introduces additional complexity and lowered the model's performance.

**DISCUSSION**

This study investigates the probabilistic prediction of LLMs in healthcare and demonstrate that the simple and flexible explicit probability (directly extracted from the generated text following the prompt) provide relatively high AUROC and AUPRC, but is consistently less reliable than the implicit probability (derived from the transition score of the predicted label token) across different languages, datasets, and prompt designs. The performance gap between explicit and implicit probabilities is especially large for small LLMs.

Probability estimation is important because the estimates are compared with thresholds set specifically for different clinical needs, reflecting expert judgment. For example, lower thresholds are used for low-risk interventions, while higher thresholds are used for invasive procedures. Incorrect estimation can lead to misinformed decisions, compromising patient safety and treatment efficacy[32,33]. Our study reminds the LLM users of the potential drawbacks of the explicit probability, particularly for small LLMs and tasks with imbalanced label distribution, underscoring the importance of validating the reliability of LLM probabilistic predictions before relying on them in clinical settings.

We show that as LLM performance declines, the reliability of explicit probabilities worsens compared to implicit ones (**Figure 2, Supplementary Table 5**). This suggests that using explicit probabilities can amplify biases inherent in LLMs, especially when their performance is poor. In the clinical domain, due to limited training data for minority languages, lack of domain knowledge, and dataset imbalances, LLMs often underperforms[34]. Most LLMs are trained mainly on English corpora[35], while many countries' medical systems use minority languages that most LLMs are not familiar with. Furthermore, due to data access and privacy concerns[34], advanced LLMs often rely on general domain datasets and lack exposure to real-world EHR notes, which limits their effectiveness in clinical settings[24,36]. Medical datasets also frequently suffer from imbalances in phenotypes, like rare diseases or serious clinical outcomes[37], and cohort identification[7]. In such clinical scenarios, the combination of LLMs' poor performance and increased biases hinders accurate interpretation of explicit probabilities.

The flaw of the explicit probability underscores the need for cautious use in clinical settings and highlights the importance of enhancing LLM explicit probability outputs by integrating implicit probabilities, which is one of the future works of this study. One promising approach is to fine-tune LLMs' explicit probability output with the supervision of implicit probability, thereby guiding models to generate more accurate probabilities, similar to the work of improving the implicit chain of thought capability of LLMs with external CoT supervision[38]. Additional approaches that improve the LLMs' capability of numerical reasoning may also be utilized to enhance the explicit probability output[29].

Recent work has shown that during LLM inference, there is a correlation between predictive uncertainty and hallucination — a behavior where LLMs generate false facts or knowledge not supported by input[39]. In clinical settings, hallucinations from LLMs can lead to incorrect diagnoses, inappropriate treatments, and undermine trust in their use. Improving explicit probability to accurately reflect LLM uncertainty could serve as a key indicator of hallucination. A primary cause of low explicit probability is insufficient knowledge, which can drive LLMs to produce hallucinated information for predictions[40,41]. Thus, low explicit probability may signal a higher risk of hallucination. We leave the detection of LLM hallucinations with enhanced explicit probability as future work, aiming to advance efforts in detecting and mitigating LLM hallucinations in healthcare.

There are several limitations in our study. Our experiments are simplified to binary classification settings to facilitate the extraction of implicit probabilities and the calculation of AUROC and AUPRC. As a result, our conclusions about binary options do not carry over to other tasks such as multiple-choice questions answering. Besides, we do not examine the probability performance using Chain of Thought (CoT) prompting, as it is challenging to extract implicit probabilities in these cases. Additionally, our study only applies to open-sourced LLMs since it is difficult to obtain the implicit probability of the tokens using proprietary LLMs. Finally, our results are primarily based on medical datasets, the generalizability to different domains needs to be confirmed by further validation studies.

This study explores the probabilistic outputs of LLMs in the medical domain, an essential aspect of LLM application in healthcare. By comparing explicit and implicit probabilities across multiple advanced LLMs and medical datasets, we have uncovered consistent discrepancies in reliability, particularly under conditions of small model size and dataset imbalance. These findings underscore the need for caution when relying on explicit probability in clinical settings, where the stakes are high, and the accuracy of probabilistic predictions is paramount. This research informs future studies focused on enhancing the trustworthiness of LLM in healthcare by improving the quality and reliability of probabilistic predictions.

**METHODS**

**Data Source**

This study is conducted on five datasets: four open-access datasets — Measuring Massive Multitask Language Understanding Clinical Knowledge (MMLU-CK)[42,43], Measuring Massive Multitask Language Understanding College Medicine (MMLU-CM)[42,43], United States Medical Licensing Examination (USMLE)[44], Mainland China Medical Licensing Examination (MCMLE)[44] — and one internal EHR dataset, Mass General Brigham - Social Determinant of Health (MGB-SDoH)[45]. The MMLU-CK, MMLU-CM, USMLE, and MCMLE datasets are publicly available through HuggingFace, where the MMLU-CK and the MMLU-CM datasets are the "clinical_knowledge" (299 questions) (https://huggingface.co/datasets/cais/mmlu/viewer/clinical_knowledge) and the "college_medicine" (200 questions) (https://huggingface.co/datasets/cais/mmlu/viewer/college_medicine) subsets of the MMLU dataset, respectively. For the USMLE and MCMLE datasets (https://huggingface.co/datasets/bigbio/med_qa ), we randomly select a 1000-question subset due to the computation resource restriction. MGB-SDoH is a private multiple-choice dataset from real-world EHR notes of the MGB healthcare system. It contains the progress notes of 200 patients and is annotated in 9 different SDoH aspects, including marital status, number of children, employment status, educational status, lifestyle factors (use of tobacco, alcohol, illicit drugs, exercise), and cohabitation status[45]. Mass General Brigham (MGB) Institutional Review Board approved the study protocol (2020P001486). Consent to participate was not deemed required for this observational investigation.

**Experiment Setting**

We select well-performing open-source LLMs released by 6 different organizations on the LLM leaderboard hosted by LMSYS[46-54]. Within each organization's models, we primarily use the large models (Qwen2-72B-Instruct, Meta-Llama-3.1-70B-Instruct, gemma-2-27b-it, Mistral-Large-Instruct-2407, Yi-1.5-34B-Chat, and Phi-3-medium-128k-instruct) for our study. The corresponding small models (Qwen2-7B-Instruct, Meta-Llama-3.1-8B-Instruct, gemma-2-9b-it, Mistral-7B-Instruct-v0.3, Yi-1.5-9B-Chat, and Phi-3-mini-128k-instruct) are used to compare the differences in terms of model sizes on the USMLE and the MGB-SDoH datasets. All LLMs used in this study are publicly available. The details and sources of the LLMs are listed in **Supplementary Table 2**.

Our experiments are conducted on an MGB server with 8 x NVIDIA H100 GPUs. AutoTokenizer and AutoModelForCausalLM modules from HuggingFace (https://huggingface.co/docs/transformers/model_doc/auto) are used to load the tokenizer and LLM, respectively. During inference time, we set *max_new_tokens* = 64 to enable fast response generation. We set *return_dict_in_generate*=True and *output_scores*=True to enable LLM output transition scores, which are a list of float numbers representing the log probability of LLM output tokens. These scores are essential to calculate the implicit probability. To calculate the standard deviation and the statistical significance, we repeat the experiment three times with *temperature* = 0, 0.3 and 0.7 respectively. Specifically, for the standard deviation, we use the np.std function from the Python numpy package (https://numpy.org/doc/stable/reference/generated/numpy.std.html) with ddof = 1 since we want the sample standard deviation. For the statistical significance, we conduct a one-sided paired t-test (whether the implicit probability is statistically better than the explicit probability) on the explicit and the implicit probability using the stats.ttest_ref function from the Python scipy package (https://docs.scipy.org/doc/scipy/reference/generated/scipy.stats.ttest_rel.html).

**Prediction and Probability Extraction**

To utilize the AUROC and AUPRC as the evaluation metrics, commonly applied in evaluating probability predictions on binary tasks, we convert each question of the datasets from the multiple-choice to binary format (i.e., question answer with two possible answer options). Specifically, for each question in these datasets, we extract its correct option and randomly select one of the other options to build the two candidate choices. To eliminate the impact of the option order to LLMs performance[55], we randomly assign 50% of the correct options as Option A and the remaining 50% of the correct options as Option B. We then design a prompt (**Supplementary Table 1**) to instruct the LLMs to output their decisions, the probabilities of the decision being correct, and the corresponding explanations following a pre-defined format. Based on the pre-defined output format, we use regular expressions on the response text to extract LLMs' prediction and the corresponding explicit probability, as illustrated in **Figure 3b**. To extract the implicit probability, we extract the probability of the token that contains the model's prediction ("A" or "B"). Validation scripts are implemented to ensure that the extracted content aligns with the model predictions and falls within the two candidate options. We define a case as invalid if the validation fails and set the corresponding explicit and implicit probabilities as *np.nan* so that it is not used in the evaluation.

**Evaluation**

Accuracy, AUROC, and AUPRC are used as the evaluation metrics. All experiments are run three times with different LLM temperatures (0, 0.3, and 0.7), and the average and standard deviation of these metrics are calculated and reported across the three runs. AUROC is the primary metric as it is the most common metric for the probability prediction. AUPRC is used to compare the performance difference on imbalanced data experiments. Distributions from explicit and implicit probabilities are also visualized for better comparison. P-values are reported to indicate the statistically significance of the one-sided paired t-test.

**Imbalanced Datasets Analysis**

Labels in clinical practice often have imbalanced distributions, and the performance of AI models are affected by such imbalance. To investigate the difference between explicit and implicit probabilities under imbalanced datasets, we reconstruct the most commonly used medical LLM evaluation dataset, USMLE, with different imbalanced distributions. Specifically, we set a variety of ratios (5%, 10%, 30%, 50%, 70%, and 90%) of option A being the correct option. Then we randomly swap option pairs of the original dataset until the ratio of option A being correct reaches the designated values. This operation does not change the wording of the question or the options. It only changes the order of option A and B so that the ratio of option A being correct can be adjusted. Since the metric of AUPRC is a commonly used metric to evaluate model performance on datasets with imbalanced labels[56,57], we calculate the AUPRC of the two models and compare their differences .

**Sensitivity Analysis**

To ensure the robustness of our results, we conduct a sensitivity analysis by using two additional prompts (**Supplementary Table 1**) on the USMLE dataset and analyze if the relative difference between the explicit and the implicit probability persists under different prompt settings. Except the prompt, all experiment settings are the same with the main experiment.

## DATA AVAILABILITY

The MMLU-CK, MMLU-CM, USMLE, and the MCMLE datasets are publicly available. The MGB-SDoH dataset is available from Mass General Brigham (MGB) but restrictions apply to the availability of these data, which were used under license for the current study to protect patient privacy, and so are not publicly available. Data are however available from the authors upon reasonable request and with permission of Mass General Brigham (MGB).

## CODE AVAILABILITY

The code implementation of this study is made publicly available at https://github.com/gubowen2/Probabilistic-Medical-Predictions-of-Large-Language-Models.

## ACKNOWLEDGEMENTS

This study was funded by the National Institute of Health (R01LM013204). The funders had no role in the design, collection, analysis, interpretation of the data, or the decision to submit the manuscript for publication.

## AUTHOR CONTRIBUTIONS

B.G. conducted the experiments and analysis and drafted the manuscript. R.D. provided the proprietary data of the study. J.K.L supervised the study. J.Y. designed the study, drafted the manuscript, and supervised the study. All authors revised, read, and approved the manuscript. B.G. takes responsibility for the integrity of the work.

## COMPETING INTERESTS

The authors declare no competing interests.

**FIGURES**

**Figure 1: AUROCs of the large and small LLMs.**

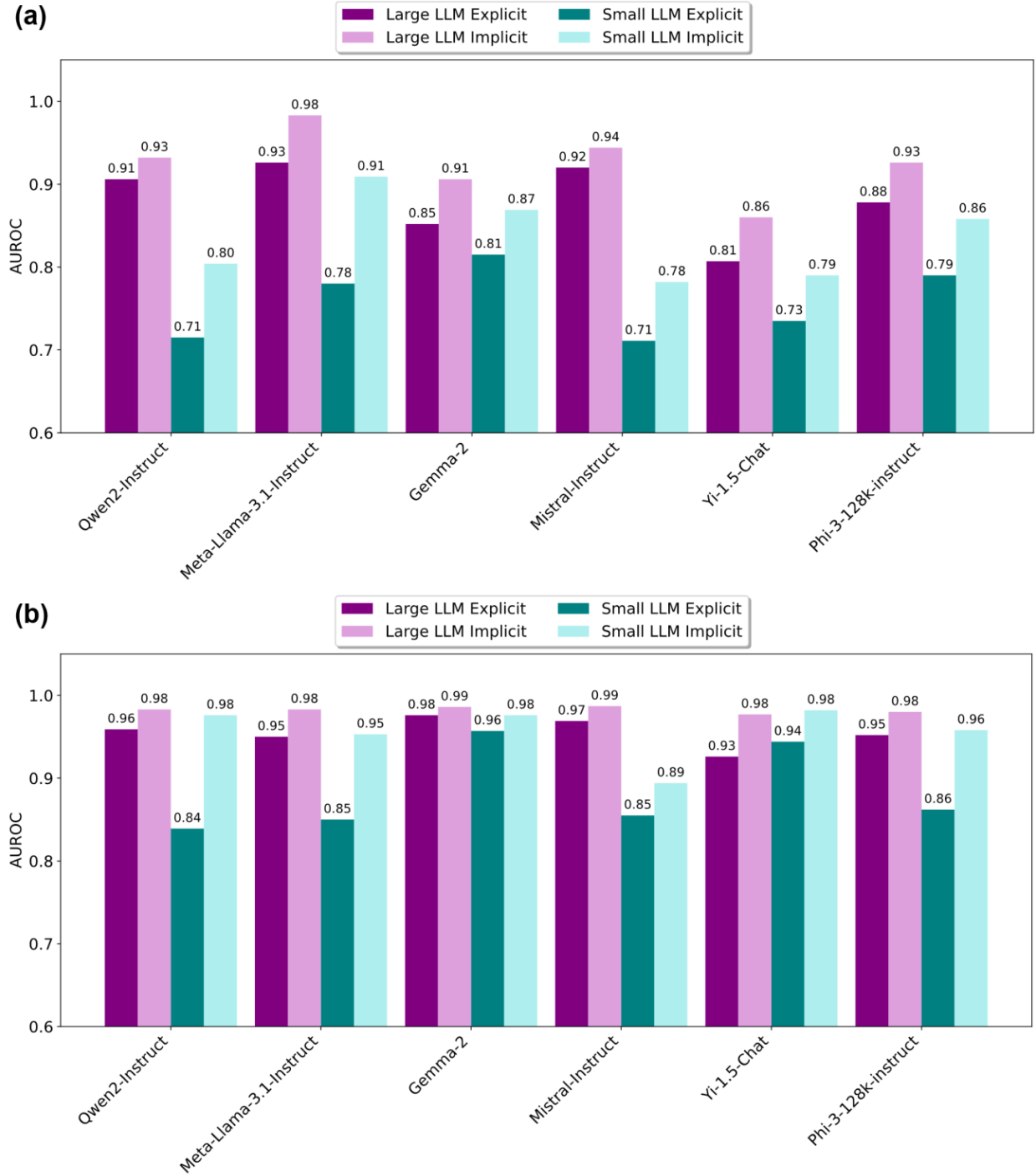

AUROCs of the large and small LLMs. **a** AUROCs of the large and small LLMs on the USMLE dataset. **b** AUROCs of the large and small LLMs on the MGB-SDoH dataset.

**Figure 2: AUPRC difference and probability distribution on the USMLE dataset.**

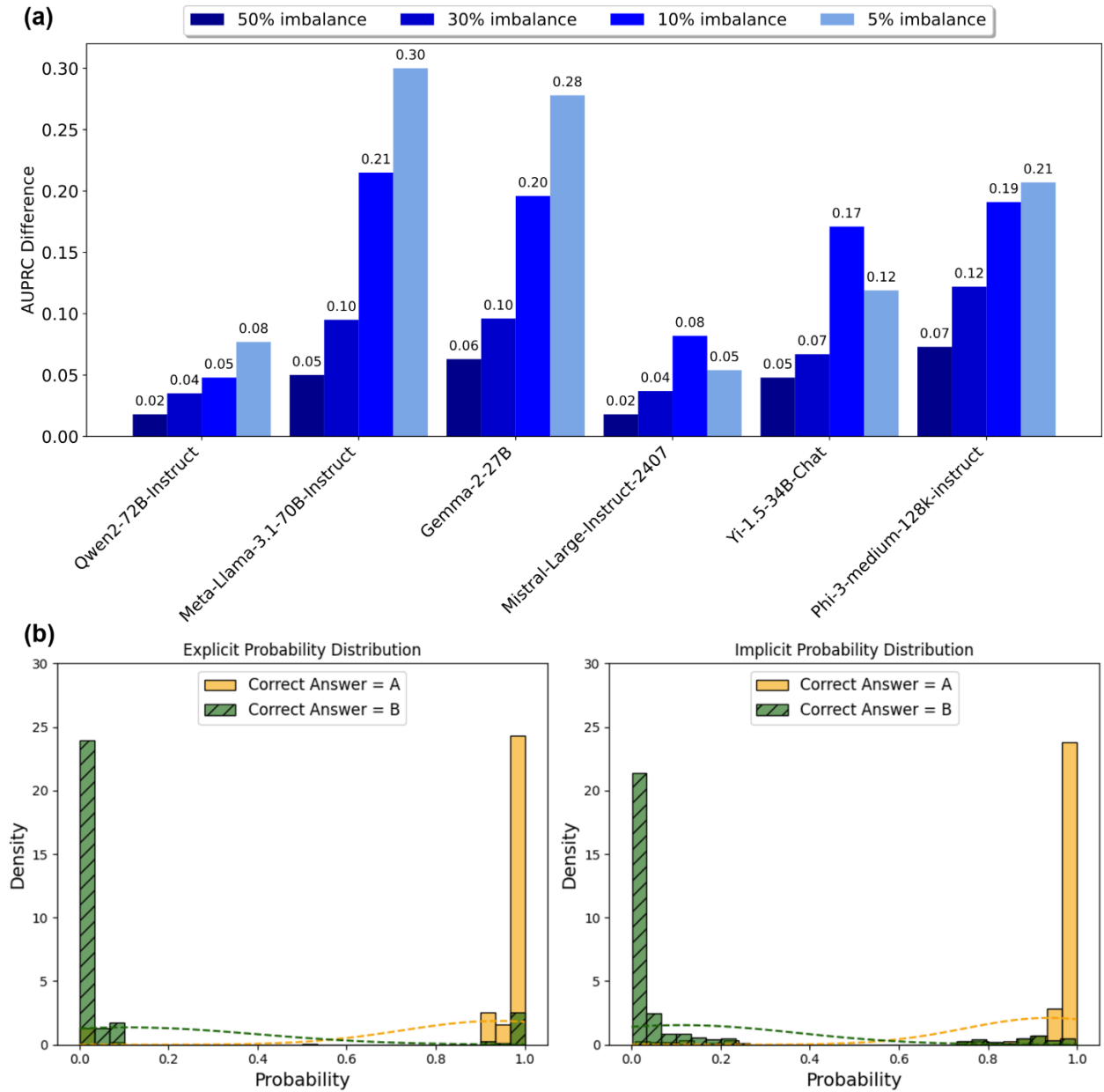

AUPRC difference and probability distribution on the USMLE dataset. **a** AUPRC difference of the large LLMs on the imbalanced USMLE dataset. **b** Probability distribution for Meta-Llama-3.1-70B-Instruct on the USMLE dataset. The dashed lines are the Gaussian distribution fitting curves for the cases when the correct answer is option A (yellow) and option B (green), respectively.

**Figure 3: Comparison between probabilistic predictions of AI models and study design.**

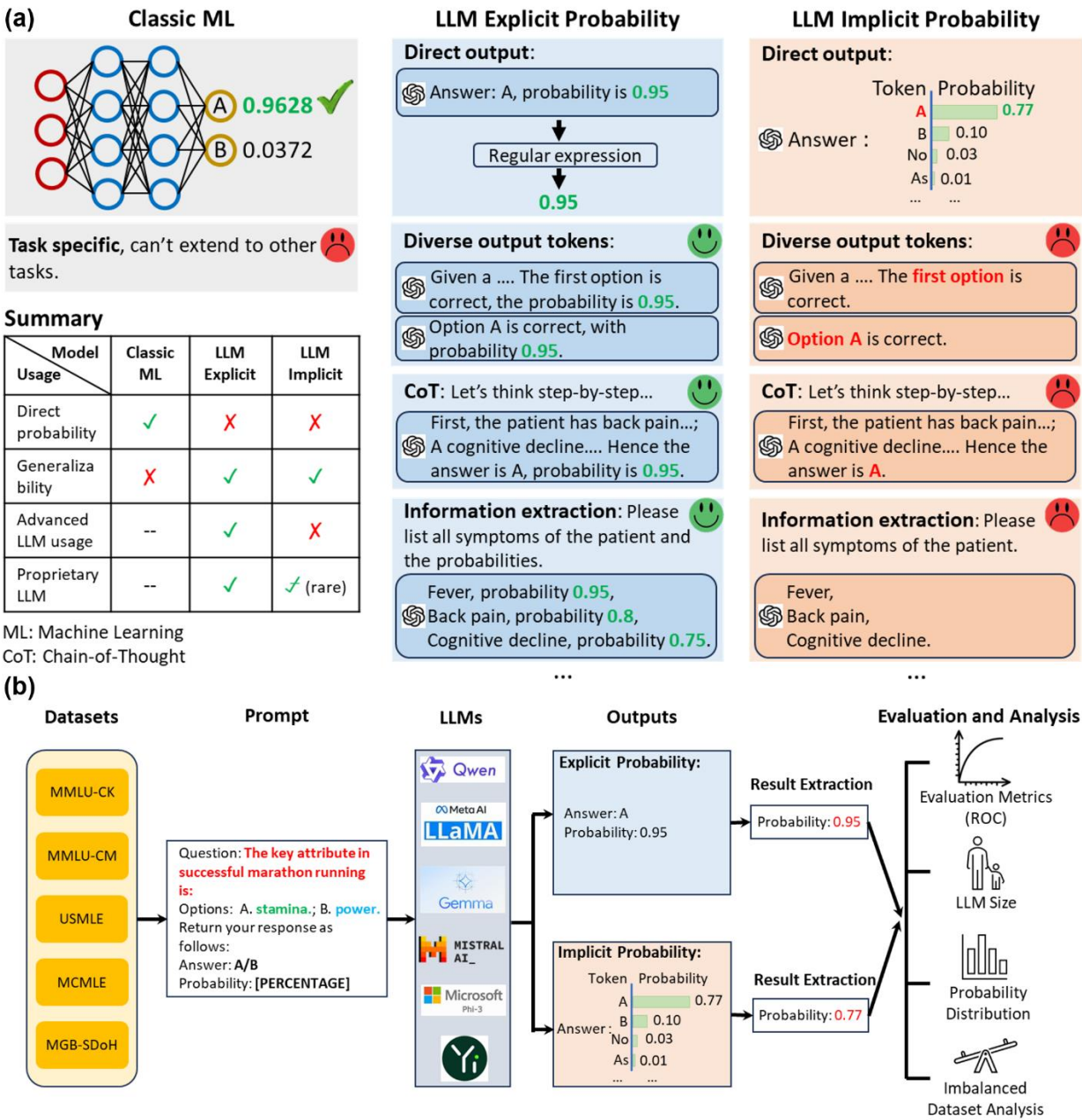

Comparison between probabilistic predictions of AI models and study design. **a** Comparison of probabilistic predictions from different types of AI models. The green happy face means the model applies to the use case, while the red unhappy face means it doesn't. **b** The conceptual framework of this study.

**TABLES**

**Table 1. Accuracy of the LLMs on different datasets.**

| Model | Average (%) | MMLU-CK | MMLU-CM | USMLE | MCMLE | MGB-SDoH |
|---|---|---|---|---|---|---|
| gemma-2-27b-it | 86.7±0.1 | 88.8±0.2 | 83.6±0.6 | *83.4±0.4* | 82.1±0.2 | 95.3±0.1 |
| Mistral-Large-Instruct-2407 | 90.0±0.5 | 91.8±0.6 | **87.6±0.3** | 89.2±0.3 | 85.2±1.2 | **96.0±0.3** |
| Yi-1.5-34B-Chat | 86.1±0.3 | *86.5±0.9* | 82.6±0.9 | 77.6±0.5 | 90.3±0.2 | *93.6±0.2* |
| Phi-3-medium-128k-instruct | *83.8±1.0* | 91.0±1.1 | *80.5±0.8* | 83.6±2.1 | *69.8±1.5* | 94.0±1.2 |
| Qwen2-72B-Instruct | 91.3±0.1 | 90.0±0.3 | 87.3±0.5 | 86.7±0.4 | **96.7±0.2** | 95.7±0.2 |
| Meta-Llama-3.1-70B-Instruct | **91.7±0.3** | **93.4±1.0** | 85.5±0.0 | **92.7±0.2** | 93.0±0.2 | 94.0±0.2 |

Boldface indicates the best performance among the 6 models under comparison in each dataset.
*Italics indicates the worst performance among the 6 models under comparison in each dataset.*
The number following the "±" symbol indicates the standard deviation.

**Table 2. AUROC of the LLMs on different datasets.**

| Model | AUROC (%) | MMLU-CK | MMLU-CM | USMLE | MCMLE | MGB-SDoH |
|---|---|---|---|---|---|---|
| gemma-2-27b-it | Explicit | 91.6±0.6 | 86.9±0.8 | 85.4±0.4 | 85.7±3.3 | 97.4±0.2 |
| | Implicit | 95.9±0.5 | 92.0±0.8 | 90.2±0.4 | 90.0±0.5 | 98.2±0.4 |
| | Difference | 4.3±0.2* | 5.1±1.2* | 4.9±0.7* | 4.2±3.7* | 0.8±0.3* |
| | Explicit | 93.2±0.3 | 89.7±0.5 | 91.8±0.5 | 87.2±3.3 | 97.1±0.3 |

| | | MMLU-CK | MMLU-CM | USMLE | MCMLE | MGB-SDoH |
|---|---|---|---|---|---|---|
| Mistral-Large-Instruct-2407 | Implicit | 96.6±0.4 | 94.6±0.6 | 94.1±0.3 | 91.0±0.8 | 98.4±0.3 |
| | Difference | 3.4±0.2* | 4.8±0.5* | 2.4±0.6* | 3.8±3.3 | 1.3±0.6* |
| Yi-1.5-34B-Chat | Explicit | 90.0±1.1 | 85.6±0.7 | 80.6±0.2 | 92.8±2.2 | 92.8±0.3 |
| | Implicit | 94.8±0.9 | 90.8±0.6 | 85.4±0.8 | 96.3±0.8 | 97.5±0.4 |
| | Difference | 4.8±2.0* | 5.2±1.3* | 4.8±0.6* | 3.5±3.0 | 4.7±0.4* |
| Phi-3-medium-128k-instruct | Explicit | 93.4±0.6 | 83.7±1.7 | 86.1±1.7 | 75.8±4.2 | 94.4±0.8 |
| | Implicit | 96.8±0.2 | 91.8±0.5 | 92.2±0.5 | 80.5±0.9 | 97.9±0.1 |
| | Difference | 3.4±0.6* | 8.1±2.2* | 6.1±1.2* | 4.7±4.1 | 3.5±0.7* |
| Qwen2-72B-Instruct | Explicit | 92.8±0.8 | 88.9±0.8 | 90.5±0.2 | 97.1±0.3 | 95.9±0.2 |
| | Implicit | 93.1±3.5 | 90.9±3.3 | 89.5±3.2 | 97.7±1.1 | 97.0±1.2 |
| | Difference | 0.3±3.2 | 2.0±3.3 | -0.9±3.1 | 0.6±0.8 | 1.1±1.2 |
| Meta-Llama-3.1-70B-Instruct | Explicit | 93.8±0.4 | 86.9±1.1 | 93.2±0.6 | 94.0±1.8 | 95.1±0.2 |
| | Implicit | 97.7±1.6 | 92.2±2.7 | 96.3±2.0 | 97.4±1.4 | 96.6±1.5 |
| | Difference | 4.0±1.4* | 5.3±3.3 | 3.1±2.3 | 3.4±0.3 | 1.4±1.6 |

*p-value < 0.05.
The number following the "±" symbol indicates the standard deviation.

**Table 3. AUPRC of the LLMs on different datasets.**

| Model | AUPRC (%) | MMLU-CK | MMLU-CM | USMLE | MCMLE | MGB-SDoH |
|---|---|---|---|---|---|---|
| gemma-2-27b-it | Explicit | 91.8±0.3 | 87.8±0.4 | 83.0±0.6 | 84.4±0.2 | 97.7±0.2 |
| | Implicit | 96.1±0.5 | 92.4±0.9 | 89.4±0.8 | 89.9±0.9 | 97.9±0.3 |
| | Difference | 4.3±0.3* | 4.6±1.3* | 6.4±0.5* | 5.5±1.0* | 0.1±0.1 |
| Mistral-Large-Instruct-2407 | Explicit | 94.7±0.4 | 92.3±0.4 | 91.9±0.6 | 87.5±0.9 | 97.6±0.3 |
| | Implicit | 97.3±0.2 | 95.4±0.5 | 93.6±0.1 | 90.3±0.7 | 98.0±0.2 |
| | Difference | 2.6±0.2* | 3.1±0.2* | 1.7±0.6* | 2.7±0.8* | 0.4±0.2* |
| Yi-1.5-34B-Chat | Explicit | 88.6±1.2 | 84.0±0.9 | 80.8±0.2 | 92.7±1.0 | 93.6±0.3 |

|  | Implicit | 94.8±0.2 | 91.7±0.8 | 85.0±0.3 | 96.8±0.5 | 97.9±0.2 |
|---|---|---|---|---|---|---|
|  | Difference | 6.1±1.2* | 7.6±1.7* | 4.2±0.5* | 4.1±1.4* | 4.3±0.3* |
| Phi-3-medium-128k-instruct | Explicit | 93.1±0.5 | 82.5±2.4 | 82.9±1.9 | 72.0±1.7 | 95.6±0.9 |
|  | Implicit | 95.3±1.0 | 91.3±0.5 | 90.2±0.4 | 80.3±0.7 | 96.5±0.9 |
|  | Difference | 2.2±1.2* | 8.8±2.8* | 7.3±1.6* | 8.3±2.0* | 1.0±0.3* |
| Qwen2-72B-Instruct | Explicit | 93.9±0.5 | 90.3±0.5 | 89.9±0.6 | 97.3±02. | 97.1±0.1 |
|  | Implicit | 94.1±1.4 | 92.4±1.6 | 90.3±0.6 | 98.1±0.6 | 97.7±0.5 |
|  | Difference | 0.2±1.0 | 2.1±1.6 | 0.4±0.8 | 0.8±0.5 | 0.5±0.5 |
| Meta-Llama-3.1-70B-Instruct | Explicit | 95.2±0.4 | 90.1±0.8 | 93.2±0.2 | 93.8±0.6 | 96.4±0.1 |
|  | Implicit | 98.1±1.1 | 93.7±2.0 | 96.6±1.7 | 97.6±1.3 | 97.2±0.9 |
|  | Difference | 2.9±0.9* | 3.6±2.3 | 3.4±1.7* | 3.8±1.5* | 0.8±0.9 |

*p-value < 0.05.
The number following the "±" symbol indicates the standard deviation.